\documentclass[]{spie}  

\usepackage{amsmath,amsfonts,amssymb}
\usepackage{graphicx}
\usepackage{amsmath}
\usepackage{graphicx}
\usepackage{amsmath}
\usepackage{amssymb}
\usepackage{xspace}
\usepackage{algorithmic}
\usepackage{enumitem}
\usepackage{multirow}
\usepackage{graphicx}
\usepackage{multirow}
\usepackage[table,xcdraw]{xcolor}
\usepackage[colorlinks=true, allcolors=blue]{hyperref}

\title{AdaMixup: A Dynamic Defense Framework for Membership Inference Attack Mitigation}

\author[a]{Ying Chen}
\author[b]{Jiajing Chen}
\author[c]{Yijie Weng}
\author[d]{ChiaHua Chang}
\author[e]{Dezhi Yu}
\author{Guanbiao Lin*}
\affil[a]{University of Connecticut}
\affil[b]{Courant Institute, New York University}
\affil[c]{Robert H. Smith School of Business, University of Maryland College Park, College Park}
\affil[d]{Metropolitan College, Department of Computer science, 
Boston University}
\affil[e]{School of Information, University of California}

\begin{document} 
\maketitle

\begin{abstract}
Membership inference attacks have emerged as a significant privacy concern in the training of deep learning models, where attackers can infer whether a data point was part of the training set based on the model's outputs. To address this challenge, we propose a novel defense mechanism, AdaMixup. AdaMixup employs adaptive mixup techniques to enhance the model's robustness against membership inference attacks by dynamically adjusting the mixup strategy during training. This method not only improves the model's privacy protection but also maintains high performance. Experimental results across multiple datasets demonstrate that AdaMixup significantly reduces the risk of membership inference attacks while achieving a favorable trade-off between defensive efficiency and model accuracy. This research provides an effective solution for data privacy protection and lays the groundwork for future advancements in mixup training methods.
\end{abstract}

\keywords{Membership Inference Attack Defense, Deep Learning, Dynamic Strategy, AdaMixup.}

\section{INTRODUCTION}
\label{sec:intro}  

As deep learning models become ubiquitous across domains such as healthcare\cite{abdullah2022review}, finance\cite{culkin2017machine}, and autonomous systems\cite{ferreira2021comparison}, concerns regarding the privacy of training data have intensified\cite{yu2024advanced}. A particularly troubling vulnerability is the susceptibility of these models to membership inference attacks (MIA)\cite{salem2018ml,shokri2017membership}. In such attacks, an adversary seeks to determine whether a specific data instance was part of the training set by analyzing the model's outputs. This vulnerability is exacerbated by the tendency of deep learning models to overfit to training data, especially when dealing with sensitive information. Consequently, MIAs pose a serious threat to data privacy in machine learning systems.

The core issue driving membership inference attacks lies in the disparity between a model's performance on training data versus data unseen during training. Deep neural networks, especially when over-parameterized, often demonstrate high confidence on training samples, which inadvertently provides attackers with exploitable signals to differentiate between members and non-members of the training set. As a result, mitigating overfitting has become a key focus in defending against MIAs. A range of defenses have been proposed to address this challenge, with techniques such as differential privacy\cite{li2021membership}, regularization\cite{ying2020privacy,doi:10.1142/S0218001424500204}, and Memguard\cite{jia2019memguard} being the most prominent. However, many of these methods come with inherent trade-offs. Differential privacy, for example, introduces noise to protect data privacy, often at the cost of model utility. 

To overcome limitations, mixup\cite{ji2023mixup} training improves model generalization and reduces overfitting. Traditional mixup uses a fixed ratio, which may not be optimal. We propose AdaMixup, which adapts the mixup ratio based on model performance. AdaMixup balances accuracy and privacy protection, enhancing generalization and MIA defense.

The main contributions of our work can be summarized as follows:
\begin{itemize}[label=$\bullet$, left=2em]
    \item We introduce AdaMixup, a novel defense method that dynamically adjusts the mixup ratio during training, enhancing model robustness against membership inference attacks.
    \item We achieve a balance between defending against membership inference attacks and maintaining high model accuracy with AdaMixup. 

\end{itemize}

\section{Background}

In this section,  we introduce various existing types of membership inference attacks. Finally, we summarize various existing state-of-the-art defense methods.

\subsection{Existing Membership Inference Attacks}

A membership inference attack aims to determine if a data point was used to train a specific machine learning model, threatening individual privacy. Due to growing concerns, many attack techniques have emerged. This paper categorizes these attacks, which are also key to the defense strategies discussed.

\subsubsection{Attacks Based on Confidence Scores.} An ubiquitous manifestation of Membership Inference Attack (MIA) is the confidence-driven approach, initially unveiled by Shokri et al.\cite{shokri2017membership} in their groundbreaking 2017 research, and then further extended by Salem et al.\cite{salem2018ml}. This assault modality exploits the confidence ratings (equivalent to prediction probabilities) furnished by the model to deduce membership status.

Formally, Prediction confidence corresponding to training samples $F(\mathbf{x})_y$ is typically higher
than prediction confidence for testing samples. Therefore, confidence-based attack will only regard the queried sample as a member when the prediction confidence is larger than either a class-dependent threshold $\tau_y$ or a class-independent threshold $\tau$
\begin{equation}
I_{\text {conf }}(F(\mathbf{x}), y)=\mathbb{1}\left\{F(\mathbf{x})_y \geq \tau_{(y)}\right\}
\end{equation}

\subsubsection{Attacks Based on Labels.} In 2021, Choquette et al.\cite{choquette2021label} introduced Label-only Membership Inference Attacks, exploiting model responses to input alterations. They noted models are more consistent in predicting training data labels. Attackers use this predictability to deduce data membership. By slightly perturbing inputs and observing label predictions, they calculate the probability of label consistency. Formally, given an input $x$ and its true label $y$, the adversary applies a small perturbation $\delta$ to generate $x^{\prime}=x+\delta$. The model's predicted label $\hat{y}\left(x^{\prime}\right)$  is observed, and the process is repeated across multiple perturbations. 
The probability that the label remains unchanged is calculated as: 
\begin{equation}
P\left(\hat{y}\left(x^{\prime}\right)=y\right)=\frac{1}{N} \sum_{i=1}^N \mathbf{1}\left\{\hat{y}\left(x_i^{\prime}\right)=y\right\}
\end{equation}
Where $N$ is the number of perturbations, and $\mathbf{1}\{\cdot\}$ is the indicator function.



\subsection{Existing Defense Methods}
Existing defenses against membership inference attacks include differential privacy\cite{ying2020privacy,cui2023member}, regularization techniques\cite{hu2023artificial,yang2024applyingconditionalgenerativeadversarial}, and mixup\cite{guo2019mixup} methods. However, these defenses have some drawbacks. Differential privacy, while effective in providing privacy guarantees, can often lead to a trade-off with model performance, resulting in reduced accuracy. Regularization techniques, such as $L_2$ regularization, can help mitigate overfitting, but they may not be sufficient to fully protect against sophisticated attacks. Mixup methods, which augment the training data by interpolating between samples, can improve generalization, but they may also suffer from limitations, such as using fixed interpolation ratios that may not be optimal for all datasets and training stages. Therefore, there is a need for continued research and development of more effective and efficient defenses against membership inference attacks.





\begin{figure*}[!h]
    \centering
    \includegraphics[scale=0.5]{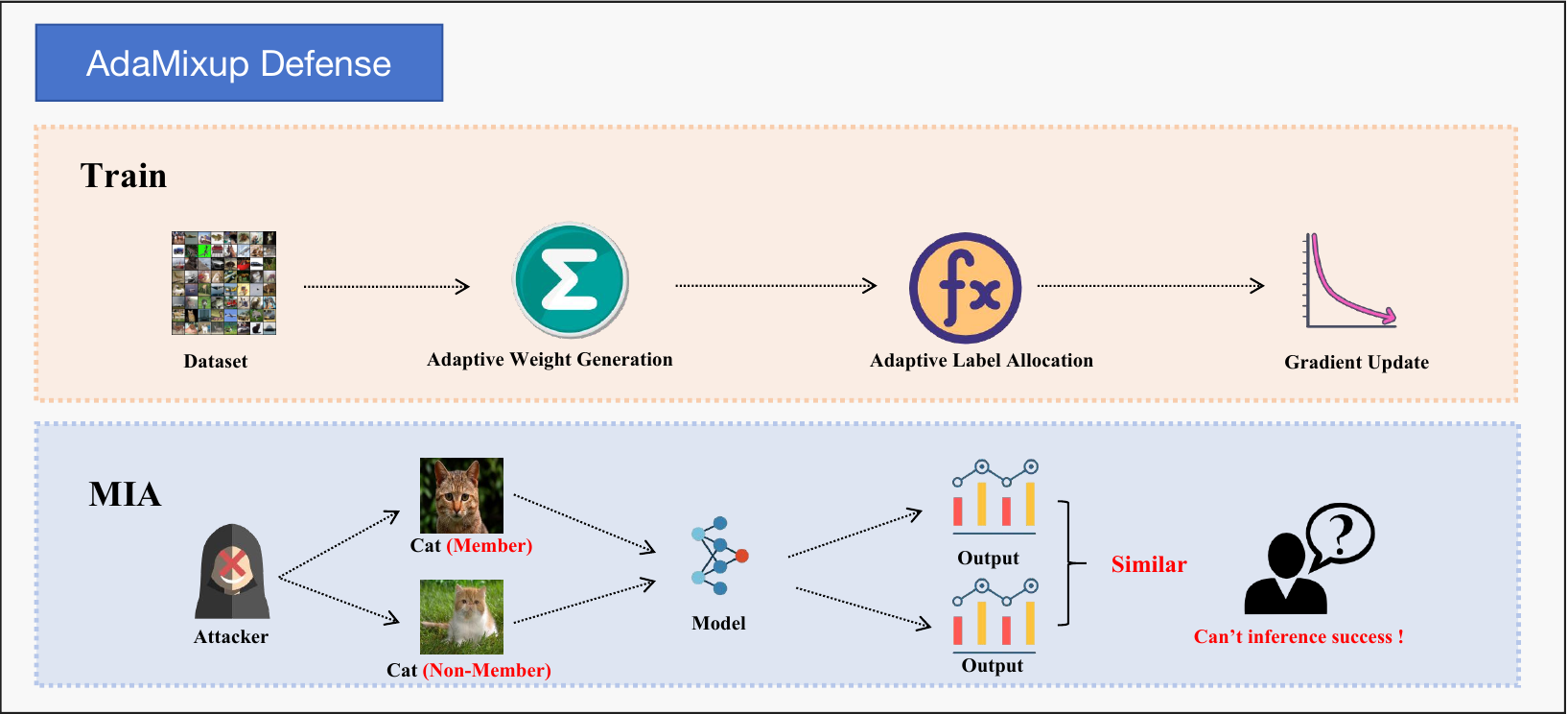} 
    \caption{Pipeline of AdaMixup Defense framework} 
    \label{fig:pipeline} 
\end{figure*}

\section{Methods}
\label{sec:misc}


In this section, we introduce AdaMixup, a defense strategy against membership inference attacks. AdaMixup integrates adaptive weight generation and label allocation, dynamically adjusting sample influence during training to boost model robustness. By contrast to traditional Mixup with a fixed $\lambda$, AdaMixup adapts $\lambda$ over epochs, preventing label distortion and improving classification performance, all while preserving high accuracy.

\begin{figure*}[!h]
    \centering
    \includegraphics[scale=0.32]{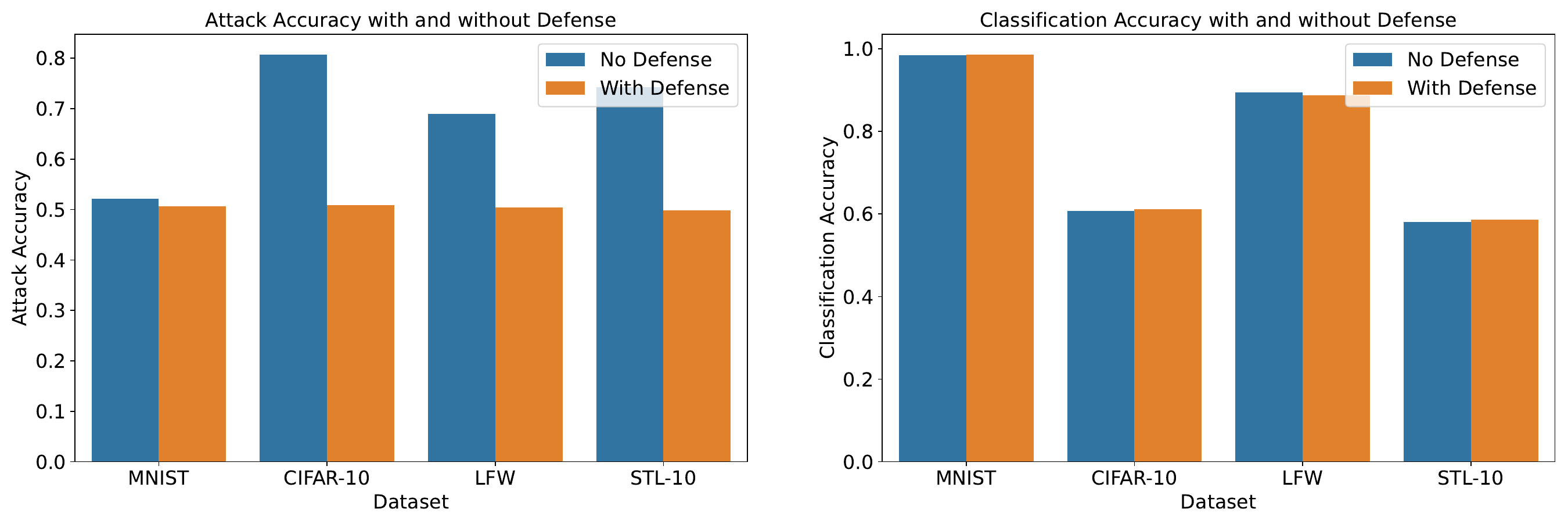} 
    \caption{Comparison of MIA attack accuracy(A1) and model classification accuracy with and without AdaMixup on different datasets.} 
    \label{fig:compare1} 
\end{figure*}

\subsection{Phase1: Adaptive Dynamic Weight Generation}
In standard Mixup, $\lambda$ is randomly sampled from a Beta distribution, which results in a uniform and uncontrolled contribution of samples throughout training. This method, while effective in regularization, may lead to suboptimal performance as the training progresses, particularly in later stages where precise classification is more critical. 

In AdaMixup, $\lambda$ is initially large to ensure substantial mixing between samples during early training. As the training process proceeds, $\lambda$ is gradually reduced, allowing the model to learn from individual, unaltered samples in later stages. This strategy balances the need for regularization in early epochs with the necessity of fine-grained learning as the model converges. We formalize the decay of 
 $\lambda$ over time as follows:
 \begin{equation}
\lambda_t=\lambda_{\text {initial }} \cdot\left(1-\frac{t}{T}\right)
\end{equation}

Where: $\lambda_t$ is the mixing coefficient at epoch $t$,  $\lambda_initial$ is the initial $\lambda$ value (typically close to 1), $T$ is the total number of training epochs.

This linear decay ensures that in the early epochs, the model is regularized by strong mixing, which helps to smooth decision boundaries and mitigate overfitting. As training progresses, $\lambda$ decreases, allowing the model to focus on learning precise representations from the original data samples. The gradual reduction in mixing intensity prevents over-regularization and ensures the model can refine decision boundaries without excessive distortion from mixed samples.

\subsection{Phase2: Adaptive Label Allocation}
A common criticism of Mixup training is the blending of labels, which can introduce ambiguity, particularly when the samples come from significantly different classes. This blending may degrade the model's performance on classification tasks, as the labels may no longer align with the semantic meaning of the original samples. To address this issue, AdaMixup incorporates an adaptive label assignment strategy, ensuring that the label assigned to the mixed sample is consistent with the sample that contributes more significantly to the mixture.

Given two samples, $x_1$ and $x_2$, with corresponding one-hot encoded labels, $y_1$ and $y_2$, the mixed sample $\tilde{x}$ is defined as:
\begin{equation}
\tilde{x}=\lambda_t \cdot x_1+\left(1-\lambda_t\right) \cdot x_2
\end{equation}
However, unlike traditional Mixup, where the mixed label $\tilde{y}$ is a weighted combination of $y_1$ and $y_2$, AdaMixup ensures that the label is determined by the dominant sample in the mixture:
\begin{equation}
\tilde{y}= \begin{cases}y_1, & \text { if } \lambda_t \geq 0.5 \\ y_2, & \text { if } \lambda_t<0.5\end{cases}
\end{equation}

This approach maintains label consistency, which is critical for classification tasks where the label must reflect the primary content of the input. By adapting the label based on the larger contribution in the mixture, AdaMixup avoids the inaccuracies introduced by label interpolation, thus improving classification performance while still benefiting from the regularization effects of Mixup.

\section{Experimental Results}
In this section, we present the experimental validation of our proposed AdaMixup method. We evaluate its effectiveness in defending against membership inference attacks and compare its performance to other established defense mechanisms.



\begin{table*}[htbp]
\setlength{\belowcaptionskip}{10pt} 
\caption{Comparison of defense results and classification accuracy of different defense methods.} 
\centering 
\label{tab:1}
\renewcommand\arraystretch{1.8}
\tabcolsep=0.2cm
\resizebox{1\linewidth}{!}{
\begin{tabular}{cc|c|c|c|c|c|c|c|c}
\hline
\multicolumn{2}{c|}{\cellcolor[HTML]{FFFFFF}}                                                                                                 & w/o   & Dropout & \begin{tabular}[c]{@{}c@{}} Mixup\end{tabular} & \begin{tabular}[c]{@{}c@{}}L1   \\ Regularization\end{tabular}    & \begin{tabular}[c]{@{}c@{}}L2   \\ Regularization\end{tabular}     & DP-SGD & MemGuard & AdaMixup \\ \hline \hline

\multicolumn{1}{c|}{\cellcolor[HTML]{FFFFFF}}                                                                                 & MNIST         & 98.84     & 98.14      & 98.01 & 98.16 & 98.07 & 94.86  & 98.84        & 98.94\\ \cline{2-10} 

\multicolumn{1}{c|}{\cellcolor[HTML]{FFFFFF}}                                                                                 & CIFAR-10       & 60.95     & 60.92    & 61.32  & 60.28 & 61.23 & 56.21   & 60.95      & 61.21  \\ \cline{2-10} 
\multicolumn{1}{c|}{\cellcolor[HTML]{FFFFFF}}                                                                                 & LFW     & 89.87    & 89.34   & 89.74   & 88.24 & 88.89 & 86.01  & 89.87    & 89.84 \\ \cline{2-10} 

\multicolumn{1}{c|}{\multirow{-5}{*}{\cellcolor[HTML]{FFFFFF}\begin{tabular}[c]{@{}c@{}}Classification\\ Accuracy\end{tabular}}}    & STL-10          & 58.76 & 58.89   & 57.31   & 58.34 & 58.46 & 53.11  & 58.76    & 58.99 \\ \hline \hline

\multicolumn{1}{c|}{\cellcolor[HTML]{FFFFFF}}                                                                                 & MNIST         & 53.72 & 50.11    & 50.06   & 50.96  & 50.14  & 50.12    & 49.92     &50.02 \\ \cline{2-10} 

\multicolumn{1}{c|}{\cellcolor[HTML]{FFFFFF}}                                                                                 & CIFAR-10       & 80.03 & 59.22   & 55.24 & 63.11  & 62.30 & 50.31   & 50.06   & 50.01 \\ \cline{2-10} 
\multicolumn{1}{c|}{\cellcolor[HTML]{FFFFFF}}                                                                                 & LFW      & 67.21 & 53.16  & 52.21 & 54.32 & 54.83 & 50.09  & 50.03    & 50.07 \\ \cline{2-10} 

\multicolumn{1}{c|}{\multirow{-5}{*}{\cellcolor[HTML]{FFFFFF}\begin{tabular}[c]{@{}c@{}}A1\\ Attack\\ Accuracy\end{tabular}}}        & STL-10          & 74.43 & 62.45    & 55.23  & 62.72  & 61.02  & 50.19  & 50.32     & 50.06  \\ \hline \hline

\multicolumn{1}{c|}{\cellcolor[HTML]{FFFFFF}}                                                                                 & MNIST         & 52.47 & 51.02      & 50.34  & 50.12    & 51.75 & 50.31   & 50.78    & 50.00 \\ \cline{2-10} 

\multicolumn{1}{c|}{\cellcolor[HTML]{FFFFFF}}                                                                                 & CIFAR-10       & 80.75 & 56.32    & 54.14  & 58.96     & 58.22  & 50.16  & 51.12     & 50.01    \\ \cline{2-10} 
\multicolumn{1}{c|}{\cellcolor[HTML]{FFFFFF}}                                                                                 & LFW      & 65.42  & 54.47   & 52.92  & 55.31 & 55.26 & 50.32  & 50.87     & 49.97 \\ \cline{2-10}

\multicolumn{1}{c|}{\multirow{-5}{*}{\cellcolor[HTML]{FFFFFF}\begin{tabular}[c]{@{}c@{}}A2\\ Attack\\ Accuracy\end{tabular}}} & STL-10          &70.21    & 56.32   & 53.13  & 54.03 & 54.12 & 50.16     & 50.43     & 50.09   \\ \hline \hline

\multicolumn{1}{c|}{\cellcolor[HTML]{FFFFFF}}                                                                                 & MNIST         & 51.23 & 50.33      & 50.94  & 50.15    & 50.47 & 49.89   & 51.23    & 50.06 \\ \cline{2-10} 

\multicolumn{1}{c|}{\cellcolor[HTML]{FFFFFF}}                                                                                 & CIFAR-10       & 78.43 & 54.41    & 53.16  & 54.79     & 54.23  & 50.21  & 81.94     & 50.04    \\ \cline{2-10} 
\multicolumn{1}{c|}{\cellcolor[HTML]{FFFFFF}}                                                                                 & LFW      & 64.37  & 54.17   & 54.22  & 55.09 & 54.26 & 50.07  & 64.37     & 49.97 \\ \cline{2-10} 

\multicolumn{1}{c|}{\multirow{-5}{*}{\cellcolor[HTML]{FFFFFF}\begin{tabular}[c]{@{}c@{}}A3\\ Attack\\ Accuracy\end{tabular}}} & STL-10          & 71.79    & 55.21   & 54.40  & 55.78 & 55.82 & 51.09     & 71.79     & 50.14   \\ \hline \hline

\end{tabular}}
\end{table*}

\subsection{Datasets and Experimental Details}
In this paper, we conducted experiments on four datasets: MNIST, CIFAR-10, LFW, and STL-10 to verify the superiority of our method. We conducted experiments with consistent settings, using a batch size of 128, a learning rate of 0.001 with Adam, and 100/50 epochs for CIFAR-10/STL-10 and MNIST/LFW. The AdaMixup mixup ratio decayed from 1.0 to 0.1. Each experiment was repeated five times. We focused on defending against two confidence-based attacks (A1, A2) by Salem et al.\cite{salem2018ml}and one label-based attack (A3) by Choquette et al.\cite{choquette2021label}, providing a comprehensive evaluation.

\subsection{Defense Methods for Comparison}
To evaluate AdaMixup, we compared it with Differential Privacy ($\epsilon$=1.0), Dropout (rate=0.5), L1/L2 Regularization, Memguard, and standard Mixup. Each defense has unique strategies to mitigate membership inference attacks.

\subsection{Analysis of Results}

We tested AdaMixup on MNIST, CIFAR-10, LFW, and STL-10 datasets. Results in Figure\ref{fig:compare1} and Table\ref{tab:1}  show AdaMixup significantly reduces attack accuracy compared to no defense. For instance, CIFAR-10's attack accuracy drops from 80.03$\%$ to 50.01$\%$ with AdaMixup. Similar trends are seen on other datasets, proving AdaMixup's effectiveness against membership inference attacks.

Table \ref{tab:1} provides a comprehensive comparison of classification accuracy and attack accuracy across different defense strategies. It is clear from the table that AdaMixup not only effectively defends against attacks but also maintains high classification performance. For instance, in the MNIST dataset, AdaMixup achieves a classification accuracy of 98.94$\%$, which is almost identical to the no-defense scenario's accuracy of 98.84$\%$. This indicates that AdaMixup introduces minimal performance degradation while providing strong protection. Similarly, for CIFAR-10, AdaMixup achieves a classification accuracy of 61.21$\%$, which is on par with the baseline (60.95$\%$), while significantly outperforming other defense mechanisms such as Differential Privacy (56.21$\%$) and L1 regularization (60.28$\%$).

Additionally, the results from the LFW and STL-10 datasets further confirm the robustness of AdaMixup. For instance, on the LFW dataset, the classification accuracy remains high at 89.84$\%$, with a substantial reduction in attack accuracy from 67.21$\%$ (without defense) to 50.07$\%$ (with AdaMixup). Similarly, for STL-10, AdaMixup maintains a classification accuracy of 58.99$\%$, while reducing the attack accuracy to 50.06$\%$, compared to 74.43$\%$ without any defense.

In summary, the results shown in both Figure \ref{fig:compare1} and Table \ref{tab:1} demonstrate that AdaMixup achieves a favorable balance between model performance and defense effectiveness. Across all four datasets, it consistently reduces attack accuracy while preserving classification accuracy. Compared to traditional defense methods such as Differential Privacy, Dropout, and regularization techniques, AdaMixup shows superior performance, especially in scenarios where preserving both accuracy and privacy is critical. This highlights the potential of AdaMixup as a robust defense mechanism against membership inference attacks.

\section{Conclusion}

This paper introduces AdaMixup, an innovative adaptive defense mechanism that dynamically adjusts mixup ratios during training to effectively mitigate membership inference attacks while preserving model accuracy. Through rigorous experiments on diverse datasets such as CIFAR-10, MNIST, LFW, and STL-10, AdaMixup has demonstrated significant reductions in attack success rates, outperforming other widely-used defense strategies. By balancing privacy defense and model performance, AdaMixup offers a promising solution for real-world applications where both privacy and performance are critical.




\bibliography{report} 
\bibliographystyle{spiebib} 

\end{document}